\pdfoutput=1

\documentclass[11pt]{article}

\usepackage{booktabs}
\usepackage[table]{xcolor}
\usepackage{svg}
\usepackage{tcolorbox}
\newcommand{\openai}{\raisebox{-0.1em}{\includegraphics[width=1em]{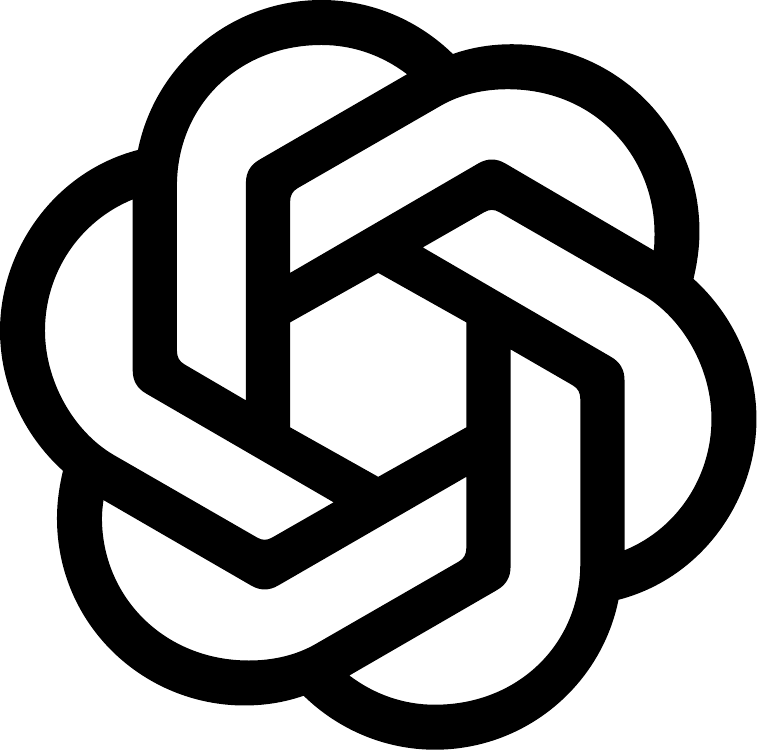}}}
\newcommand{\deepseek}{\raisebox{-0.1em}{\includegraphics[width=1em]{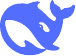}}}
\newcommand{\qwen}{\raisebox{-0.1em}{\includegraphics[width=1em]{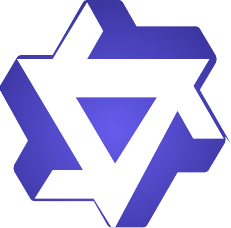}}}
\newcommand{\meta}{\raisebox{0em}{\includegraphics[width=1em]{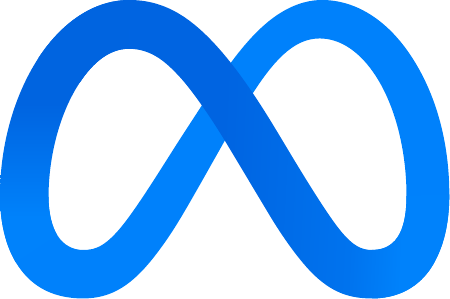}}}

\tcbuselibrary{listings, skins, breakable, theorems}

\newtcolorbox[auto counter]{responsebox}[2][]{%
  float=htb,
  title={Box~\thetcbcounter: #2},
  label={},
  colback=bluebg,
  colframe=blue!60!black,
  coltitle=white,
  fonttitle=\bfseries,
  fontupper=\small\ttfamily,
  boxrule=0.5mm,
  rounded corners,
  width=\textwidth,
  enhanced,
  breakable,
  #1 
}

\newtcolorbox[auto counter]{responsebox*}[2][]{%
  float*=htb,
  title={Box~\thetcbcounter: #2},
  label={},
  colback=bluebg,
  colframe=blue!60!black,
  coltitle=white,
  fonttitle=\bfseries,
  fontupper=\small\ttfamily,
  boxrule=0.5mm,
  rounded corners,
  width=\textwidth,
  enhanced,
  breakable,
  #1
}

\expandafter\def\csname tcb@cnt@responsebox*autorefname\endcsname{Box}

\newcommand{\pdfbox}[1]{
  \begin{minipage}{0.45\linewidth}
    \centering
    \includegraphics[width=\linewidth]{#1}
  \end{minipage}
}

\usepackage[final]{acl}

\usepackage{times}
\usepackage{latexsym}

\usepackage[T1]{fontenc}

\usepackage[utf8]{inputenc}
\usepackage{adjustbox}

\usepackage{microtype}

\usepackage{inconsolata}

\usepackage{graphicx}
\definecolor{answer_attempt}{HTML}{2c7bb6}
\definecolor{hedge}{HTML}{abd9e9}
\definecolor{clarification}{HTML}{fdae61}
\definecolor{refuse}{HTML}{d7191c}

\definecolor{gptfour}{HTML}{018571}
\definecolor{qwen}{HTML}{80cdc1}     
\definecolor{deepseek}{HTML}{a6611a}     
\definecolor{gptmini}{HTML}{dfc27d}
\definecolor{llama}{HTML}{9467bd}
\definecolor{dpollama}{HTML}{8c564b}

\definecolor{bluebg}{HTML}{E2ECF6}

\title{It Depends: Resolving Referential Ambiguity in Minimal Contexts with Commonsense Knowledge}

\author{Lukas Ellinger \and Georg Groh \\
  School for Computation, Information and Technology \\
  Technical University of Munich, Germany \\
  \texttt{\href{mailto:lukas.ellinger@tum.de}{lukas.ellinger@tum.de}, grohg@cit.tum.de}}

\usepackage{enumitem}
\usepackage{makecell}

\begin{document}
\maketitle
\begin{abstract}
Ambiguous words or underspecified references require interlocutors to resolve them, often by relying on shared context and commonsense knowledge. Therefore, we systematically investigate whether Large Language Models (LLMs) can leverage commonsense to resolve referential ambiguity in multi-turn conversations and analyze their behavior when ambiguity persists. Further, we study how requests for simplified language affect this capacity. Using a novel multilingual evaluation dataset, we test DeepSeek v3, GPT-4o, Qwen3-32B, GPT-4o-mini, and Llama-3.1-8B via LLM-as-Judge and human annotations. Our findings indicate that current LLMs struggle to resolve ambiguity effectively: they tend to commit to a single interpretation or cover all possible references, rather than hedging or seeking clarification. This limitation becomes more pronounced under simplification prompts, which drastically reduce the use of commonsense reasoning and diverse response strategies. Fine-tuning Llama-3.1-8B with Direct Preference Optimization substantially improves ambiguity resolution across all request types. These results underscore the need for advanced fine-tuning to improve LLMs' handling of ambiguity and to ensure robust performance across diverse communication styles.
\end{abstract}

\section{Introduction}
\begin{figure}[t]
    \includegraphics[width=\linewidth]{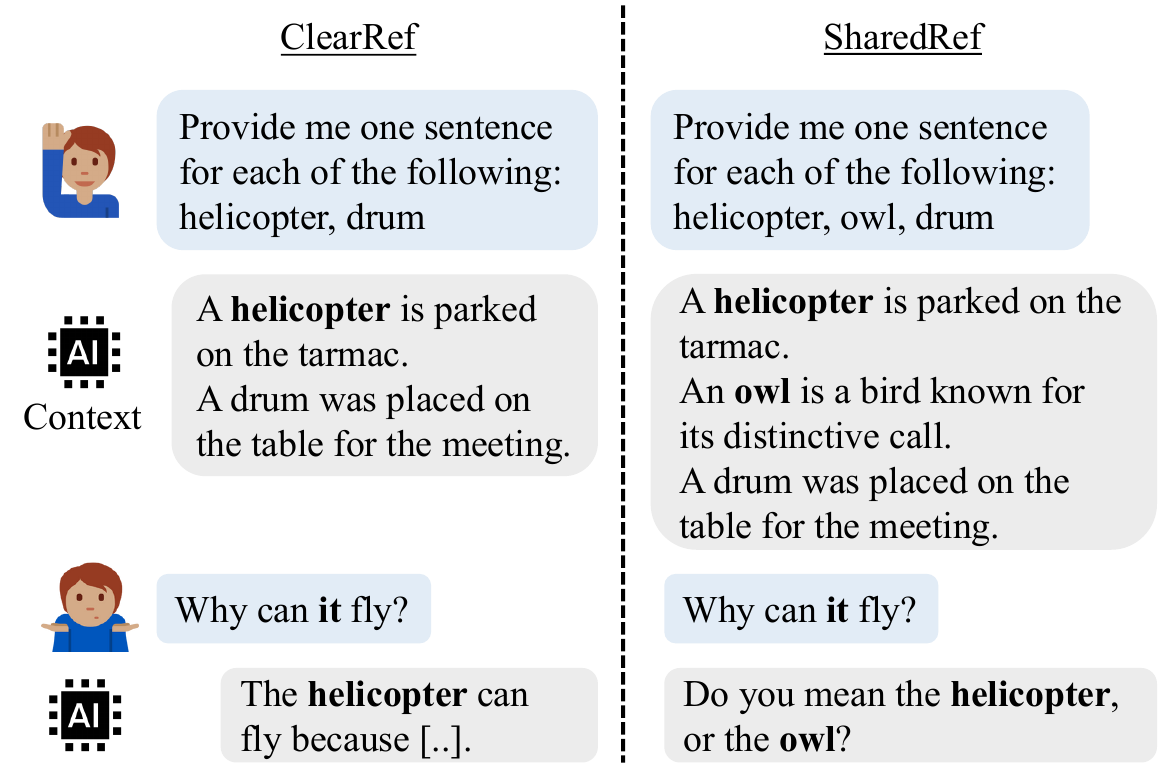}
    \caption{Two conversations between a user and an LLM in response to the ambiguous question (``Why can \textbf{it} fly?''). In both cases, the LLM uses prior context to narrow the possible referents to entities capable of flying. In the left conversation, it attempts an answer; in the right, it asks for clarification.}
    \label{fig:abstract}
\end{figure}

Natural language is inherently ambiguous. For example, pronouns may refer to multiple possible entities within a sentence. Nevertheless, humans typically resolve such ambiguity by drawing on context, shared knowledge, and conversational history \citep{ferreira_ambiguity_2008}. Consider the two conversations shown in \autoref{fig:abstract}, where the user asks the question, ``Why can \textbf{it} fly?''. Without additional clues, the pronoun ``it'' is unclear and could refer to multiple entities. In the left conversation, the prior context mentions a helicopter and a drum; in the right, it additionally includes an owl. Humans effortlessly combine this context with commonsense knowledge, recognizing that drums cannot fly, but helicopters and owls can. As a result, the first case is unambiguous, while the second may require clarification.

This process reflects a fundamental feature of human communication: a ``division of labor'' between speakers and listeners, where speakers omit explicit details to minimize effort, trusting listeners to fill in gaps using common ground \citep{ferreira_ambiguity_2008}. Common ground consists of the mutual knowledge, beliefs, and assumptions interlocutors accumulate and maintain during conversation \citep{clark_grounding_1991, clark_using_1996}. Central to common ground is commonsense knowledge, a broadly shared understanding of the world that enables people to make implicit inferences effortlessly.

As mentioned, humans are usually good at building and using common ground. While prior work suggests that LLMs struggle with ambiguity resolution, particularly in static, single-turn contexts \citep{liu_were_2023}, our work shifts focus to a conversational setting. We study how LLMs behave in multi-turn dialogs where common ground is explicitly established through conversation history and commonsense knowledge. In our setting, multiple referents can remain plausible even after considering prior context. This allows us to evaluate how models handle uncertainty through different response strategies, such as requesting clarification. 

We further examine how language constraints affect this ability. Language models are increasingly used to generate output in different variants, such as simplified and easy-to-understand language. This has clear benefits for accessibility, particularly for users with cognitive or linguistic challenges \citep{freyer_easy-read_2024}. However, simplified outputs often reduce the depth and precision of content \citep{trienes_infolossqa_2024}. \citet{ellinger_simplifications_2025} find that models prompted to define homonyms in simple language often default to the most salient meaning, disregarding less dominant but valid definitions. We explore whether such requests for simplified language also affect a model’s capacity to resolve ambiguity when multiple interpretations are plausible.

Studying this is crucial because misinterpretation of ambiguous language can lead to downstream failures such as misinformation, hallucinations, or user confusion. By systematically testing whether LLMs consider multiple plausible candidates rather than relying on recency or default biases, we provide a diagnostic view of their behavior in ambiguous conversational settings.

Our contributions are as follows:
\begin{itemize}
    \item We introduce a multilingual dataset for evaluating LLMs to resolve referential ambiguity in conversations with explicit common ground.
    \item We evaluate DeepSeek v3, GPT-4o, Qwen3-32B, GPT-4o mini, and Llama 3.1 8B using both LLM-as-Judge and human annotations.
    \item We show that LLMs often commit to a single interpretation or cover all references instead of hedging or clarifying. Simplified language constraints worsen this by reducing commonsense reasoning and response diversity.
    \item We fine-tune LLaMA 3.1 8B with Direct Preference Optimization (DPO), achieving significant improvements on our task that generalize to a lexical ambiguity benchmark, with less degradation under simplified prompts.
\end{itemize}

\section{Background and Related Work}
\begin{figure*}[t]
    \includegraphics[width=\linewidth]{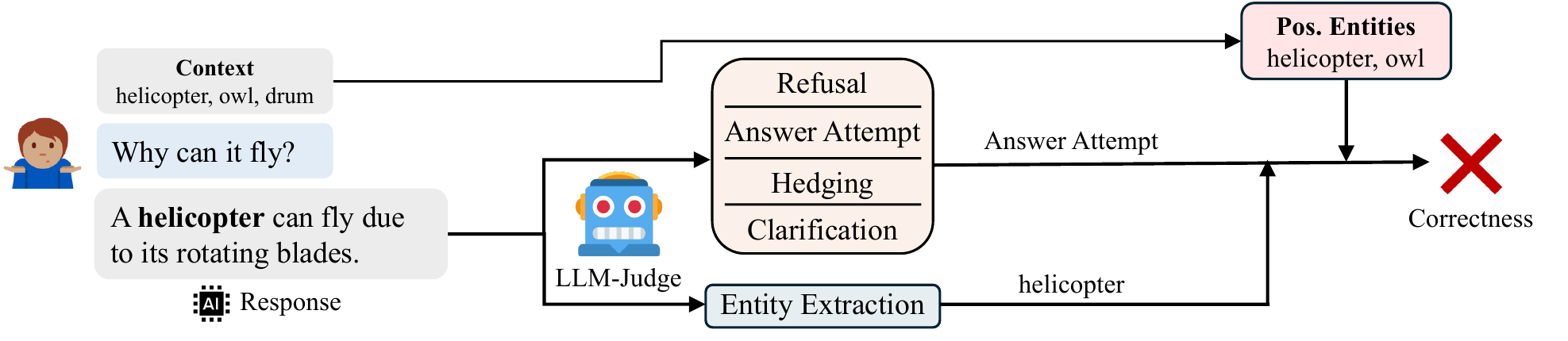}
    \caption{
    Evaluation pipeline including LLM-Judge for response categorization and entity extraction.  
    Based on these outputs and the positive entities identified in the context, we determine the fine-grained response category and assess correctness with respect to entity resolution.
    }
    \label{fig:pipeline}
\end{figure*}
\paragraph{Ambiguity and Clarification.}
Understanding language often requires resolving ambiguity, such as referential ambiguity, where it is unclear which entity a phrase refers to. Such unclear references slow down human processing \citep{gernsbacher_mechanisms_1989, macdonald_measuring_1990, myers_accessing_1998, stewart_shallow_2007}, yet humans are usually good at resolving them by drawing on common ground. 

In contrast, LLMs struggle with ambiguity. \citet{min_ambigqa_2020} introduce AmbigQA, a dataset designed to investigate underspecified questions, and subsequent studies \citep{wildenburg_pre-trained_2024, liu_were_2023} show that even state-of-the-art models underperform in such settings. This limitation extends to the multimodal domain: \citet{testoni_racquet_2024} find that vision–language models also handle ambiguity poorly, often replying with overconfident or biased outputs. While their focus is on visual context, the challenge is related to ours, with textual context instead of images.

Models also rarely seek clarification. \citet{kuhn_clam_2023} show that LLMs often respond incorrectly to ambiguous inputs rather than asking follow-up questions. Prior work confirms this lack of clarification behavior \citep{benotti_modeling_2017, xu_asking_2019, shi_learning_2022}. \citet{herlihy_overcoming_2024} link this tendency to fine-tuning biases and propose a taxonomy of model responses, which we adopt.

Prior work mainly studies ambiguity in static, single-turn settings without common ground. Notably, datasets for anaphora resolution, such as the Winograd Schema Challenge \citep{levesque_winograd_2012}, focus on single-sentence coreference, where exactly one antecedent is correct and can be identified using commonsense reasoning. In contrast, we study LLMs in multi-turn dialogs where common ground is explicitly established through conversation history and commonsense knowledge. In our setting, multiple referents can remain plausible even after considering context. This allows us to evaluate how models handle uncertainty through different response strategies, such as direct answers, hedging, or requesting clarification,  rather than simply selecting the correct noun.

Finally, we test if our fine-tuned model generalizes to lexical ambiguity using the benchmark of \citet{ellinger_simplifications_2025}, which evaluates homonym definitions without disambiguating context.

\paragraph{Commonsense Evaluation.}
Prior work systematically evaluated LLMs on commonsense reasoning benchmarks. \citet{li_systematic_2022} conduct evaluations under zero- and few-shot settings across four benchmarks, revealing that pre-trained LMs struggle to acquire commonsense knowledge without task-specific supervision. Scaling model size or adopting to few-shot prompting does not suffice to reach human-level performance. Similarly, \citet{bian_chatgpt_2024} assess ChatGPT on eleven commonsense QA datasets. They find that ChatGPT can retrieve relevant knowledge via prompting. However, it often fails to identify and apply the specific commonsense required to answer a given question. In the multimodal domain, \citet{fu_commonsense-t2i_2024} introduce Commonsense‑T2I, the first benchmark evaluating whether text‑to‑image models generate images consistent with commonsense knowledge. They find that state-of-the-art models achieve only 49\% accuracy, indicating significant gaps in visual commonsense understanding.

Our work extends these by exploring another dimension of commonsense. Unlike prior benchmarks focused on question answering or image alignment, we assess whether models recognize ambiguous referents and either disambiguate or request clarification, demonstrating a context-aware application of commonsense reasoning.

\paragraph{Simple Language.}
Simplified language aims to improve accessibility for a broad range of users, including non-native speakers, children, domain novices, and individuals with cognitive impairments. Its availability is endorsed by the Web Content Accessibility Guidelines (WCAG) to promote inclusive communication \citep{w3c_web_2025}. Simplified language involves straightforward vocabulary, clear sentence structure, minimal jargon, and the avoidance of complex grammar \citep{freyer_easy-read_2024}. Domains like healthcare, law, and education already widely apply it \citep{garimella_text_2022, deilen_towards_2024, rets_approaches_2022}. However, prior work has shown that simplification in LLM-generated text can lead to undesirable side effects such as omissions or overly vague formulations \citep{anschutz_disimproved_2025, agrawal_text_2024, devaraj_evaluating_2022}. \citet{ellinger_simplifications_2025}, for instance, report that when asked to define homonyms in simplified language, models tend to default to the most salient meaning, neglecting valid but less frequent senses.

Building on this line of work, we study how simplification constraints affect a model’s ability to resolve referential ambiguity and how task-specific finetuning affects performance in the lexical ambiguity benchmark of \citet{ellinger_simplifications_2025}.

\section{Methodology}
We evaluate whether LLMs can resolve referential ambiguity using common knowledge and how requests for simplified language affect this ability. Each test instance consists of a short context passage introducing some entities (e.g., \textit{helicopter}, \textit{owl}, \textit{drum}). The user then asks an ambiguous question referring to one of the entities without naming it directly (e.g., \textit{Why can it fly?}). For each instance, we define a set of positive entities as those for which the question makes sense, and negatives as those for which it does not (e.g., a drum cannot fly). We evaluate two setups: \textit{ClearRef}, where one positive and one negative entity make the referent unambiguous with commonsense, and \textit{SharedRef}, where two positives and one negative leave ambiguity even with commonsense. This setup tests whether models consider multiple plausible candidates rather than relying on recency or default biases. We treat the pronoun ``it'' as equally applicable to all introduced positive entities. To assess the impact of recency, we perform an ablation in which the order of entities is permuted (see \autoref{app:abl-permutation}).

\subsection{Dataset}
We construct our datasets based on ConceptNet~\citep{speer_conceptnet_2017}, a knowledge graph that encodes commonsense relationships between entities and attributes. We select eight relations, such as \textit{capable of flying}, and extract all associated entities. \autoref{fig:abmiguous-questions} provides the complete list of relations. Since each dialog requires a context passage, we use GPT-4.1-nano to generate a concise sentence for every entity. These sentences, each beginning with the entity name, serve as the context passages for all related evaluations.

For \textit{ClearRef}, each entity is paired with a negative sample from a different relation. We use GPT-4.1-nano to verify that the negative entity does not satisfy the target relation.
For \textit{SharedRef}, we create samples by pairing all entities within the same relation and similarly pick a negative. This results in 52 \textit{ClearRef} and 227 \textit{SharedRef} examples. We list further details in \autoref{app:dataset}.

To enable multilingual evaluation, we translate the context sentences and entities into Arabic, French, Russian, and Simplified Chinese using the DeepL API\footnote{\url{https://www.deepl.com/en/pro-api}}. We choose these languages to facilitate comparison with the multilingual setting of \citet{ellinger_simplifications_2025}.

\subsection{Model and Prompt Configuration}
\begin{figure}[t]
\begin{tcolorbox}[
colback=brown!10,
colframe=brown!50!black,
title=Ambiguous Questions by Relation,
fonttitle=\bfseries, 
rounded corners, 
boxrule=0.8pt]
Rel. 1: Why can it \textbf{fly}?\\
Rel. 2: Why is it \textbf{sweet}?\\
Rel. 3: Why is it \textbf{made of wood}?\\
Rel. 4: Why can it \textbf{swim}?\\
Rel. 5: Why can it \textbf{run fast}?\\
Rel. 6: Why can it \textbf{climb trees}?\\
Rel. 7: Why is it \textbf{hot}?\\
Rel. 8: Why is it \textbf{loud}?\\[0.2em]
\textbf{Simple:} [..] Respond in simple language.
\end{tcolorbox}
\caption{Ambiguous questions for our eight relations. In the Simple setting, an instruction is appended. Exact relations names in \autoref{app:dataset}.}
\label{fig:abmiguous-questions}
\end{figure}
We evaluate five LLMs on our task: GPT-4o, GPT-4o-mini~\citep{openai_gpt-4_2024}, Qwen3-32B ~\citep{qwen_team_qwen3_2025}, DeepSeek v3~\citep{deepseek-ai_deepseek-v3_2025}, and Llama 3.1 8B \citep{grattafiori_llama_2024}. These models vary in size and openness, enabling a comprehensive analysis of performance across diverse LLMs. Details on model versioning and access are listed in \autoref{app:model_access}.

We evaluate eight relations, each associated with an ambiguous question. 
For each, we test two prompt settings: 
\textbf{Normal}, presenting only the ambiguous question, and 
\textbf{Simple}, which adds an instruction to respond in simplified language. 
This setup allows us to examine how constraining outputs to simpler language affects model responses. 
English prompts are shown in \autoref{fig:abmiguous-questions}, with multilingual versions in Appendix \autoref{fig:multilingual-prompts}.

\subsection{Evaluation Pipeline}
The input to the evaluation pipeline (\autoref{fig:pipeline}) consists of a brief dialogue between a user and an LLM, exemplified in \autoref{fig:abstract}. The response to the dialogue is passed to our LLM-Judge, which performs two tasks. First, it classifies the response type into one of four categories: \textit{Refusal}, \textit{Answer Attempt}, \textit{Hedging}, or \textit{Clarification} (cf. \autoref{sec:response-categorization}). In this case, the response is labeled as an \textit{Answer Attempt}. Second, it extracts all entities mentioned in the response (here, \textit{helicopter}). Using the set of mentioned entities and the known positive entities (in this case, \textit{helicopter} and \textit{owl}), we assess the correctness of the response. Since the model attempts an answer but only mentions one of the two positive entities, the response is marked as incorrect.

\subsection{Response Categorization}
\label{sec:response-categorization}
Following \citet{laban_llms_2025}, we adopt the response taxonomy from \citet{herlihy_overcoming_2024}, which includes \textit{Answer Attempt}, \textit{Clarification}, \textit{Interrogation}, \textit{Discussion}, \textit{Hedging}, \textit{Refusal}, and \textit{Missing}. Focusing on referential ambiguity resolution, we simplify this taxonomy by merging \textit{Interrogation} into \textit{Clarification} and \textit{Discussion} into \textit{Answer Attempt}, reducing annotation complexity. Full definitions and examples appear in \autoref{app:response-categories}. Briefly:
\begin{itemize}
    \item \textbf{Hedging}: The assistant uses conditional or speculative language (e.g., ``might be...'', ``if you meant X...'').
    \item \textbf{Clarification}: The assistant requests more information without offering interpretations or using hedging.
    \item \textbf{Answer Attempt}: The assistant clearly commits to at least one interpretation, providing a factual response without any hedging.
\end{itemize}

We define a response as \textit{correct} if it appropriately addresses the ambiguity in the input. Clarifications are always correct, as they seek additional input without committing to an interpretation. Hedging responses are considered correct, as long as they mention at least one entity. While they do not resolve the ambiguity, they acknowledge it and express uncertainty in a transparent way. In contrast, answer attempts are only deemed correct if they explicitly mention both positive entities.

\citet{herlihy_overcoming_2024} discuss the trade-off between the usefulness and cognitive cost of different response categories, approximated by response length. In our setting, we argue that the most desirable responses, regardless of the category, are those that mention all and only the positive entities. We refer to these as \textbf{direct} responses. They reflect correct disambiguation based on common knowledge while minimizing user effort through clear and concise answers, free of irrelevant distractors.

In \textit{SharedRef}, we consider any \textit{direct} response the most appropriate response. In contrast, for \textit{ClearRef}, where the ambiguity can be fully resolved, an \textit{Answer Attempt} is preferred.

\subsection{Automatic Evaluation}
\label{sec:automatic-evaluation}
\begin{figure*}[t]
    \centering
    \includegraphics[width=\linewidth]{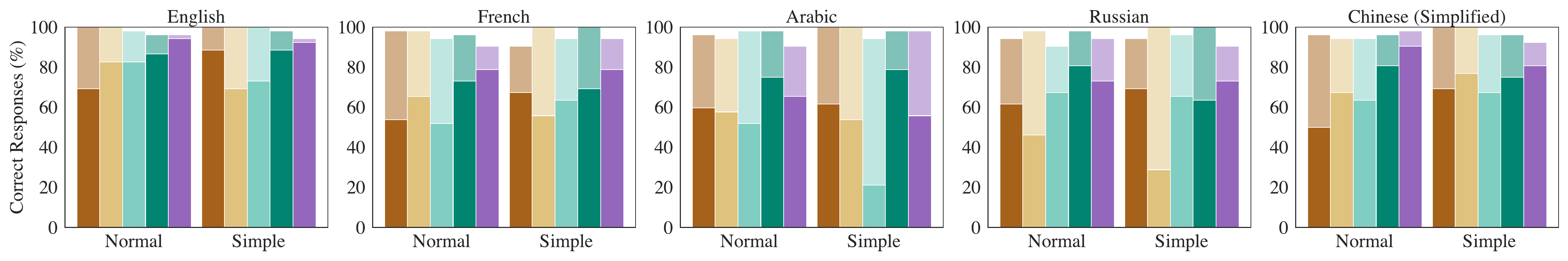}
\caption{
Percentage of correct responses across five languages on the ClearRef dataset. Colored squares indicate different models: 
\textcolor{deepseek}{\rule{1.2ex}{1.2ex}}~DeepSeek v3, 
\textcolor{gptmini}{\rule{1.2ex}{1.2ex}}~GPT-4o-mini.
\textcolor{qwen}{\rule{1.2ex}{1.2ex}}~Qwen3-32B,
\textcolor{gptfour}{\rule{1.2ex}{1.2ex}}~GPT-4o,
and \textcolor{llama}{\rule{1.2ex}{1.2ex}}~Llama-3.1-8B. The darker portion of each bar represents the percentage of Direct Responses.}
\label{fig:lang-correct-results-clearref}
\end{figure*}

\begin{figure*}[t]
    \centering
    \includegraphics[width=\linewidth]{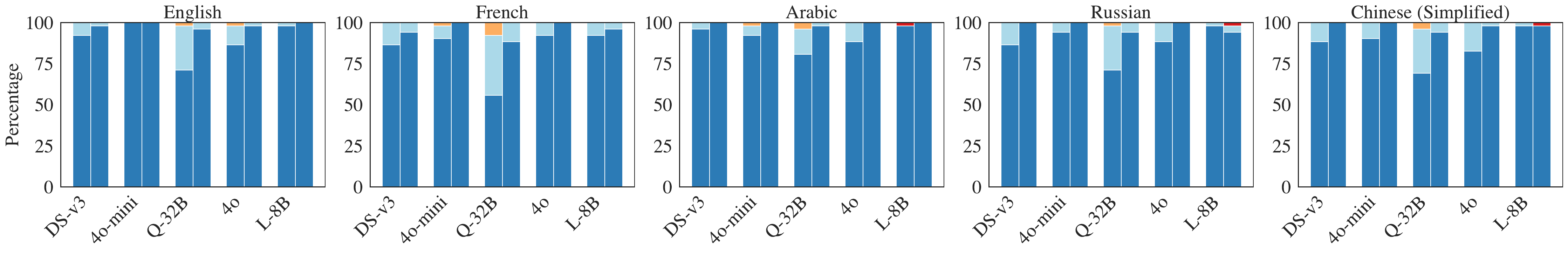}

\caption{
Distribution of the defined response categories across five languages on the ClearRef dataset. For each model, the left bar represents the Normal setting and the right bar the Simple setting.
Colored squares represent response types: 
\textcolor{answer_attempt}{\rule{1.2ex}{1.2ex}}~Answer Attempt, 
\textcolor{hedge}{\rule{1.2ex}{1.2ex}}~Hedge, 
\textcolor{clarification}{\rule{1.2ex}{1.2ex}}~Clarification, 
and \textcolor{refuse}{\rule{1.2ex}{1.2ex}}~Refuse.
}
\label{fig:lang-cat-distribution-clearref}
\end{figure*}
We designed an automated evaluation framework that leverages GPT-4.1-mini as an LLM-Judge. The framework assesses model responses based on the response categories defined in \autoref{sec:response-categorization}. It classifies responses and extracts explicitly mentioned entities. A few-shot prompt, detailed in \autoref{app:automatic-evaluation}, guides the evaluation. To validate the framework, one author manually labeled 500 responses from the English dataset, with 100 responses per evaluated model (50 for the standard prompt and 50 for the simple prompt). The annotator performed both response classification and extraction of explicitly mentioned entities, exactly as the LLM was tasked to do. The LLM judge achieved a 98\% agreement rate on response classification and a Cohen's Kappa score of 0.916, indicating almost perfect agreement according to \citet{landis_measurement_1977}. For entity extraction, the framework achieved a 97.8\% exact match accuracy. More details are provided in \autoref{app:automatic-evaluation}.

\subsection{Direct Preference Optimization}
We fine-tuned Llama-3.1-8B to improve referential ambiguity resolution using DPO~\citep{rafailov_direct_2024}. DPO aligns model behavior with desired outcomes by training on preference pairs. In our setup, we favor direct over incorrect responses.

Our training dataset contains 1,388 preference pairs across all languages by comparing incorrect Llama 3.1 8B's outputs with \textit{direct} responses from other models. To prevent reliance on entity position, we randomly permuted the order within each conversation. We restricted the training data to the \textit{`capableOf fly'} relation, allowing us to later assess generalization to other relations.

We performed a single training run using the whole training set. This decision reflects our aim to demonstrate the feasibility of aligning models to produce more useful responses with lower cognitive cost, rather than optimizing for peak performance through extensive tuning. Detailed training information is provided in \autoref{app:dpo}.  

\section{Results}
\subsection{ClearRef Dataset}
\begin{figure*}[t]
    \centering
    \includegraphics[width=\linewidth]{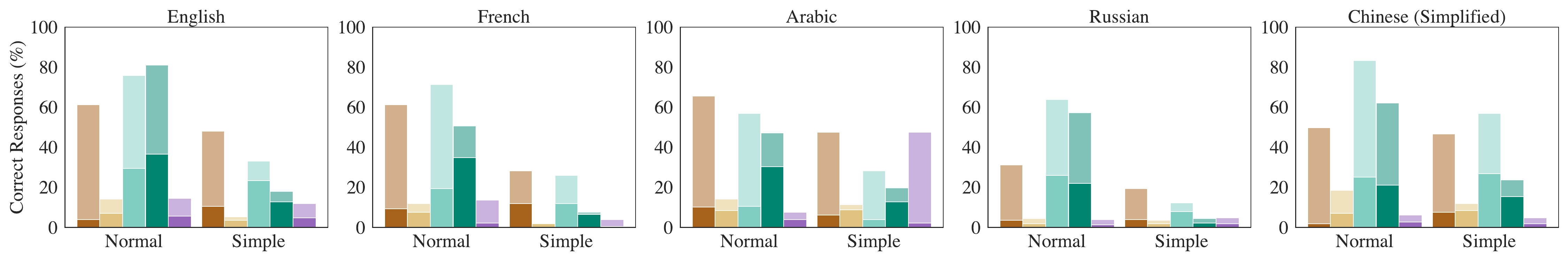}
\caption{
Percentage of correct responses across five languages on the SharedRef dataset. Colored squares indicate different models: 
\textcolor{deepseek}{\rule{1.2ex}{1.2ex}}~DeepSeek v3, 
\textcolor{gptmini}{\rule{1.2ex}{1.2ex}}~GPT-4o-mini.
\textcolor{qwen}{\rule{1.2ex}{1.2ex}}~Qwen3-32B,
\textcolor{gptfour}{\rule{1.2ex}{1.2ex}}~GPT-4o,
and \textcolor{llama}{\rule{1.2ex}{1.2ex}}~Llama-3.1-8B. 
The darker portion of each bar represents the percentage of Direct Responses.}
\label{fig:lang-correct-results}
\end{figure*}

\begin{figure*}[t]
    \centering
    \includegraphics[width=\linewidth]{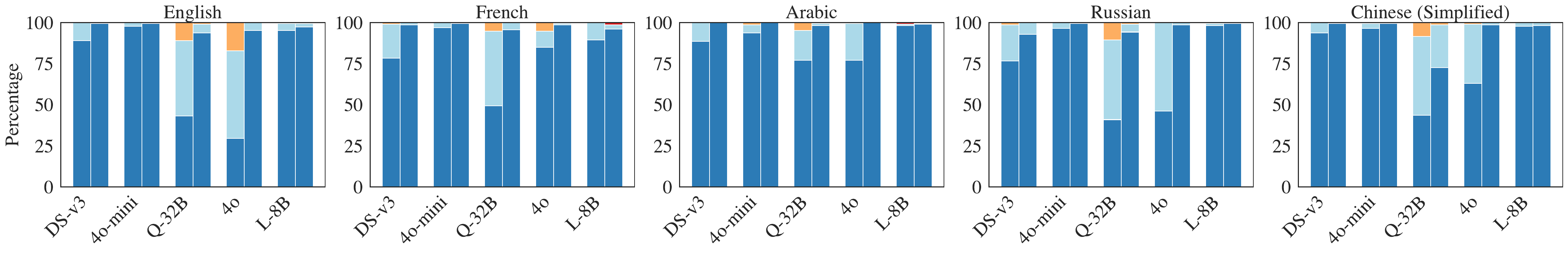}

\caption{
Distribution of the defined response categories across five languages on the SharedRef dataset. For each model, the left bar represents the Normal setting and the right bar the Simple setting. 
Colored squares represent response types: 
\textcolor{answer_attempt}{\rule{1.2ex}{1.2ex}}~Answer Attempt, 
\textcolor{hedge}{\rule{1.2ex}{1.2ex}}~Hedge, 
\textcolor{clarification}{\rule{1.2ex}{1.2ex}}~Clarification, 
and \textcolor{refuse}{\rule{1.2ex}{1.2ex}}~Refuse.
}
\label{fig:lang-cat-distribution}
\end{figure*}
\autoref{fig:lang-correct-results-clearref} shows that all models maintain correctness above 90\% across languages and settings, with some achieving perfect scores. The lowest correctness score is 90.38\%, observed for Deepseek v3 (Simple) and Llama-3.1-8B (Normal) in French. When comparing the Normal and Simple settings, GPT-4o is the only model with higher correctness in the Normal setting, while the other models either remain similar or slightly decrease. The rate of direct responses among the correct answers varies drastically across models and languages. In the Simple setting, Qwen3-32B shows the highest variance, with a direct response rate ranging from as low as 22.45\% in Arabic to 73.08\% in English. In the Normal setting, GPT-4o-mini varies most, with only 47.06\% direct responses in Russian to 82.69\% in English. Llama-3.1-8B demonstrates the highest rates for English, achieving 98.00\% in Normal and 97.96\% in Simple. Averaged across languages, mean direct responses among all responses differ by model and setting. Except for Deepseek v3, all models show higher direct response rates in the Normal setting compared to Simple. In Normal, Llama-3.1-8B achieves the highest rate (80.38\%), followed by GPT-4o, GPT-4o-mini, Qwen3-32B, and Deepseek v3 (58.85\%). Detailed breakdowns by model, language, and prompt type are provided in Appendix \autoref{tab:clear_ref}.

In \autoref{fig:lang-cat-distribution-clearref}, we show the distribution of response categories across languages and models. In all cases, \textit{Answer Attempt} is the dominant category. However, comparing the Normal and Simple settings reveals a shift: In the Simple setting, models nearly always produce answer attempts (mean 97.92\%). In Normal, especially with Qwen3-32B, hedging occurs more frequently, and to a lesser extent, clarifications. For Qwen3-32B, the average proportion of Answer Attempts drops to 69.61\%.

\subsection{SharedRef Dataset}
We show proportions of correct responses along with direct response rates in \autoref{fig:lang-correct-results}. The results reveal a sharp drop from Normal to Simple and a clear separation between two model groups: high performers (GPT-4o, Qwen3-32B, Deepseek v3) and low performers (Llama-3.1-8B, GPT-4o-mini).

Low-performing models show poor performance across languages and prompt settings, with GPT-4o-mini reaching below 13\% correctness in the Normal setting and Llama-3.1-8B slightly higher but inconsistent due to an outlier in the Arabic Simple setting.

Among the top performers, GPT-4o achieves the highest correctness in English Normal prompts (81.06\%, thereby 45.11\% direct), while Qwen3-32B performs best overall when averaged across languages in the Normal setting (70.22\%, 31.11\%). Deepseek v3 leads in the Simple setting (37.97\%, 22.73\%), outperforming the others despite lower direct response rates.

Performance also varies notably by language. In the Normal setting, English (69.16\% correct, thereby 47.28\% direct) and Chinese (63.96\%, 41.54\%) achieve the highest average correctness, followed by Arabic, French, and Russian (51.19\%, 45.71\%), reflecting the models’ native strengths (e.g., GPT for English, Qwen and Deepseek for Chinese). In the Simple setting, Arabic leads (50.22\%, 44.01\%), followed by Chinese and English, with French and Russian (26.08\%, 59.81\%) trailing. We show a detailed breakdown per model, language, and prompt type in Appendix \autoref{tab:shared_ref}.

\autoref{fig:lang-cat-distribution} shows the distribution of response categories across languages and models. Consistent with ClearRef, Answer Attempt remains the dominant category in the Simple setting, with an average proportion of 97.01\% across all languages and models. The only notable outlier is Qwen3-32B in Chinese, with a lower proportion of 72.69\%.

In the Normal setting, the shift toward other response categories becomes more pronounced than in ClearRef. The average proportion of Answer Attempts decreases to 77.67\%. Notable deviations include GPT-4o in English (29.52\%) and Russian (46.26\%), as well as Qwen3-32B in English (43.17\%), French (49.34\%), Russian (40.97\%), and Chinese (43.61\%). These two models show marked increases in Hedging (GPT-4o from 1.67\% to 35.06\%, Qwen3-32B from 8.37\% to 41.14\%) and Clarification (GPT-4o from 0.09\% to 4.76\%, Qwen3-32B from 0.70\% to 8.02\%).

\subsection{Direct Preference Optimization}
\begin{figure}[t]
    \centering

    \includegraphics[width=\linewidth]{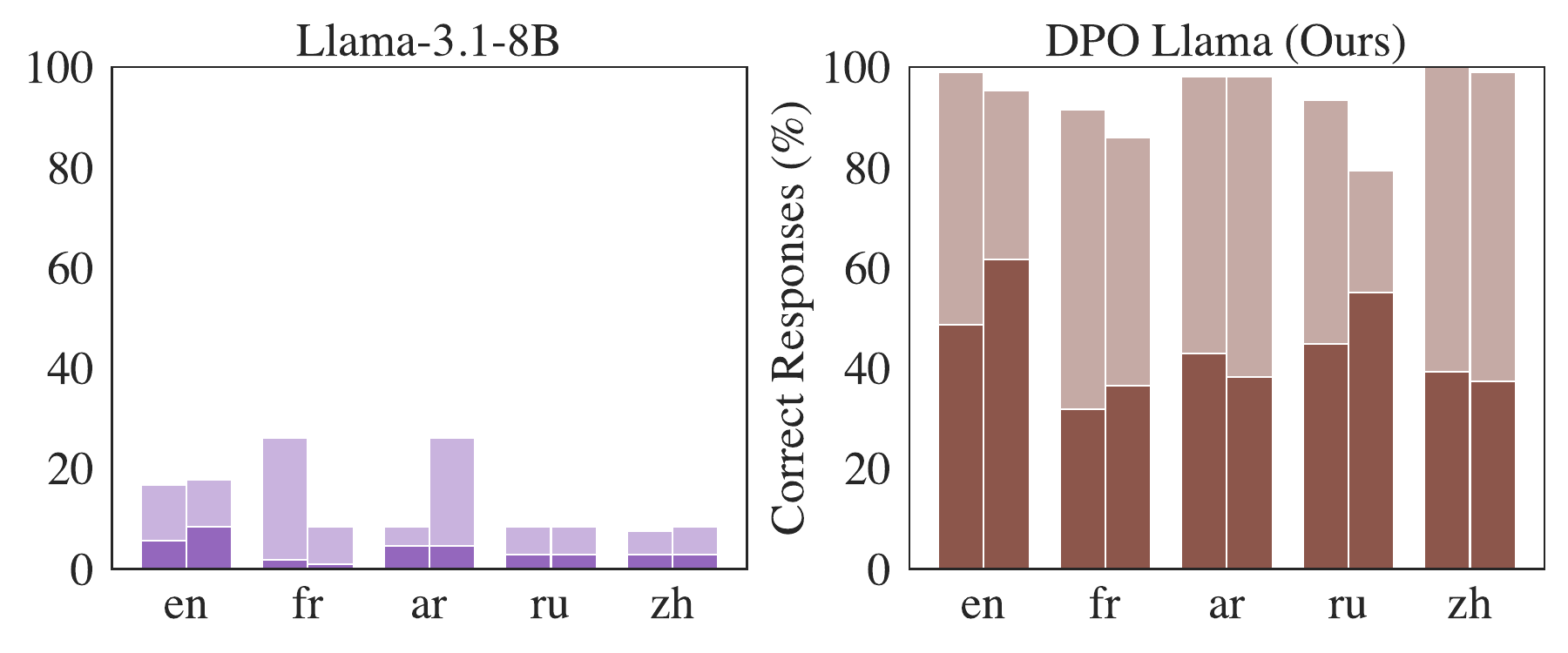}

    \includegraphics[width=\linewidth]{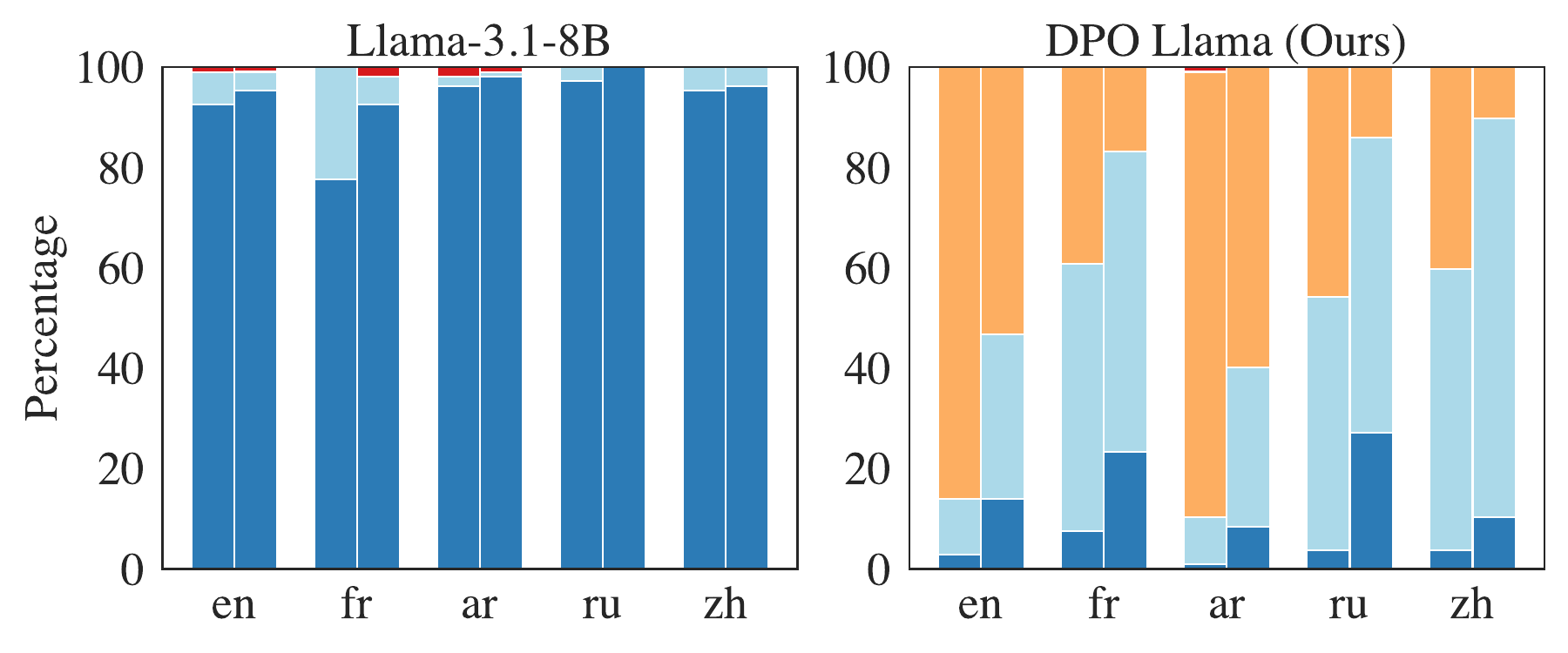}
    \caption{
    Comparison of the base model and our DPO-fine-tuned model across five languages on the SharedRef test set. 
    For each language, the left bar represents the Normal setting, and the right bar the Simple setting.\\
    Top: Percentage of correct responses. The darker portion of each bar represents the percentage of Direct Responses.\\
    Bottom: Distribution of response categories. Colored squares denote: 
    \textcolor{answer_attempt}{\rule{1.2ex}{1.2ex}}~Answer Attempt, 
    \textcolor{hedge}{\rule{1.2ex}{1.2ex}}~Hedge, 
    \textcolor{clarification}{\rule{1.2ex}{1.2ex}}~Clarification, 
    \textcolor{refuse}{\rule{1.2ex}{1.2ex}}~Refuse.
    }
    \label{fig:dpo-results}
\end{figure}

We compare the base and the fine-tuned model on the SharedRef test set, excluding the \textit{capableOf fly} relation among positives.
\autoref{fig:dpo-results} shows that the results are consistent across languages. Overall, the proportion of correct responses increases from 13.46\% to 96.45\% in the Normal setting and from 13.83\% to 91.59\% in the Simple setting. Among the correct responses, direct responses rise from 28.60\% to 42.96\% (Normal) and from 33.37\% to 50.66\% (Simple). For comparison, the best base model, Qwen3-32B, achieves 62.43\% correct (30.44\% direct) in the Normal setting and 22.06\% correct (60.97\% direct) in the Simple setting.

The category distribution shifts drastically. In the base model, Answer Attempts dominate (91.78\% in Normal, 96.45\% in Simple). After fine-tuning, Clarification is most frequent, followed by Hedging and Answer Attempts. In the Simple setting, Clarification is less dominant than in Normal, while Hedging becomes more prevalent: 60.00\% vs. 30.84\% Clarification, 36.07\% vs. 52.52\% Hedging, and 3.74\% vs. 16.63\% Answer Attempts.

\subsection{Homonym Definition Generation}
\begin{table}[t]%
\centering%
\small%
\setlength{\tabcolsep}{4pt}%
\begin{tabular}{@{}lrrrrr@{}}%
\toprule%
\textbf{Prompt / Model} & \multicolumn{5}{c}{\textbf{Sense Aware}} \\%
\cmidrule(lr){2%
-%
6}%
&En&Fr&Ar&Ru&Zh\\%
\midrule%
\multicolumn{6}{l}{\textbf{Prompt: Normal}} \\%
\meta{} 3.1 8B&\textit{96.95}&15.17&10.62&6.52&4.66\\%
\openai{} 4o-mini&93.90&79.31&92.92&90.43&84.46\\%
\qwen{} 3{-}30B A3B&94.58&86.55&\textbf{98.23}&90.00&\textbf{100.00}\\%
\meta{} 4 Maverick&96.27&54.83&74.34&75.65&45.08\\%
\deepseek{} v3&94.24&\textit{87.93}&91.15&\textit{91.74}&\textit{87.56}\\%
Our 3.1-8B&\textbf{97.63}&\textbf{97.93}&\textit{93.81}&\textbf{99.57}&84.46\\%
\color[gray]{0.4} Their 3.1-8B&\color[gray]{0.4}\text{99.66}&\color[gray]{0.4}\text{99.31}&\color[gray]{0.4}\text{99.12}&\color[gray]{0.4}\text{99.13}&\color[gray]{0.4}\text{98.45}\\%
\midrule%
\multicolumn{6}{l}{\textbf{Prompt: Simple}} \\%
\meta{} 3.1 8B&64.41&7.59&6.19&2.17&7.77\\%
\openai{} 4o-mini&63.05&52.76&\textit{76.99}&43.91&\textit{75.13}\\%
\qwen{} 3{-}30B A3B&\textbf{76.61}&\textit{59.66}&69.03&\textit{67.83}&\textbf{82.38}\\%
\meta{} 4 Maverick&\textit{69.83}&28.28&45.13&48.70&68.91\\%
\deepseek v3&63.73&47.93&\textbf{80.53}&65.22&74.09\\%
Our 3.1-8B&61.02&\textbf{73.10}&71.68&\textbf{79.13}&64.25\\%
\color[gray]{0.4} Their 3.1-8B 8B&\color[gray]{0.4}\text{92.88}&\color[gray]{0.4}\text{93.45}&\color[gray]{0.4}\text{96.46}&\color[gray]{0.4}\text{99.57}&\color[gray]{0.4}\text{94.30}\\%
\midrule%
\multicolumn{6}{l}{\textbf{Prompt: ELI5}} \\%
\meta{} 3.1 8B&7.12&7.59&0.88&1.30&0.52\\%
\openai{} 4o-mini&5.42&6.90&10.62&2.61&6.74\\%
\qwen{} 3{-}30B A3B&\textbf{22.03}&\textit{17.24}&9.73&\textit{14.78}&\textit{14.51}\\%
\meta{} 4 Maverick&10.85&13.10&11.50&9.57&9.84\\%
\deepseek v3&8.14&8.28&\textit{13.27}&8.70&10.88\\%
Our 3.1-8B&\textit{13.22}&\textbf{25.86}&\textbf{46.90}&\textbf{19.13}&\textbf{17.62}\\
\color[gray]{0.4}Their 3.1-8B&\color[gray]{0.4}\text{35.59}&\color[gray]{0.4}\text{35.17}&\color[gray]{0.4}\text{55.75}&\color[gray]{0.4}\text{63.48}&\color[gray]{0.4}\text{33.68}\\\bottomrule%
\end{tabular}%
\caption{Sense Awareness scores by prompt type and language. Best results are in \textbf{bold}, second-best in \textit{italic}. Model outputs are copied from the original paper.}
\label{tab:homonym}
\end{table}
\citet{ellinger_simplifications_2025} introduced MCL-WiC, a multilingual homonym dataset, along with the \textit{Sense Awareness} metric for evaluation. A response shows Sense Awareness by providing multiple definitions or explicitly acknowledging ambiguity via clarification requests or remarks about alternative meanings. They evaluated model performance under standard, simplified, and ELI5-style prompting \citep{fan_eli5_2019}, where the model explains a word as if the user were five years old.

\autoref{tab:homonym} compares our fine-tuned model with the results reported by \citet{ellinger_simplifications_2025}. Against baseline models, our model achieves the highest Sense Awareness under the Normal prompt in English, French, and Russian, the second-highest in Arabic, and competitive results in Chinese. For Simple, it ranks highest in French and Russian, with comparable results in other languages. For ELI5, it outperforms all baseline models in every language except English, where it ranks second. Compared to its base model, our fine-tuned version shows consistent, mostly extensive improvements across all configurations, with the only exception being the English Simple setting, where performance drops by three percentage points. 

They also fine-tuned Llama-3.1-8B on the same task. Their model produces English outputs for all languages except Russian, reflecting heavy optimization for English. In contrast, our DPO model handles all languages natively. While their fine-tuned model generally achieves higher Sense Awareness scores, our model remains competitive against the baseline models and narrows the gap in the language constraints. Their fine-tuning was explicitly targeted at this task, and reducing the gap between the language constraints. In contrast, our model achieves strong results across all languages without task-specific tuning.

\section{Discussion}
Our results indicate that current models struggle to apply commonsense knowledge for ambiguity resolution. In the simpler ClearRef task, where only one entity fits the question, models are able to resolve the ambiguity with an accuracy ranging from 94.23\% down to 21.15\% depending on the model and setting. The more challenging SharedRef task, which involves two plausible entities, sees direct responses ranging from just 36.56\% down to 0.44\%. This aligns with findings by \citet{bian_chatgpt_2024}. They observe that LLMs can retrieve commonsense facts, which in our case means realizing that an entity fits a relation when asked on its own. However, the models often fail to apply this knowledge when answering a specific question requiring such reasoning. In \autoref{app:abl-cot-prompt}, we evaluate GPT-4o’s performance in English under a Chain-of-Thought setting, prompting it to explicitly verbalize its commonsense reasoning first.

Consistent with \citet{herlihy_overcoming_2024} and \citet{kuhn_clam_2023}, we observe that models frequently skip clarification, opting to answer even when uncertainty remains. Several models show almost no clarification or hedging behavior. \citet{herlihy_overcoming_2024} and \citet{singhal_long_2024} argue that this behavior stems from reinforcement learning from human feedback (RLHF). Annotation processes typically focus on single-turn conversations. As a result, models are rarely exposed to examples of follow-up clarification questions, which require multi-turn interaction. Moreover, annotators often favor verbose, catch-all answers for under-specified queries, even though such verbosity imposes cognitive costs on users \citep{singhal_long_2024}.

Another important observation is that prompting models to use simpler language can harm response quality. Interestingly, in ClearRef, there is no drop from Normal to Simple; in some models, Simple responses are even slightly better. In contrast, for the more complex SharedRef task, performance drops drastically in the Simple setting. This confirms prior work showing that simplification often leads to omissions and vague phrasing \citep{ellinger_simplifications_2025, anschutz_disimproved_2025, trienes_infolossqa_2024, agrawal_text_2024, devaraj_evaluating_2022}. We argue that this behavior needs to change. For example, \citet{kearney_language_2025} show that LLMs adapt the information they provide based on assumptions about the user. This is problematic, especially if requesting simple language causes models to produce less thoughtful responses or overlook important distinctions. Again, RLHF may play a role, failing to capture the needs of diverse users and discouraging clarification and hedging in simplified contexts.

Taken together, we argue that resolving ambiguity requires a balance: infer as much as possible to avoid unnecessary elaboration, but clarify when uncertainty remains. Our DPO-trained model moves in this direction. It not only improves on our main evaluation but also generalizes to the lexical ambiguity benchmark of \citet{ellinger_simplifications_2025}. Moreover, it reduces the performance drop commonly observed when models operate in simplified language settings. This suggests that clarification and hedging behaviors can be learned in a transferable and robust way.

\section{Conclusion}
In this paper, we analyzed how LLMs handle textual referential ambiguity and to what extent they apply commonsense knowledge to resolve it. Our findings show that LLMs have limited ability to do so effectively. They tend to commit to a single interpretation or cover all possible references, rather than hedging or seeking clarification. This tendency becomes even more pronounced when users request simple language, which reduces commonsense reasoning and different answering strategies.

These results point to two core issues. First, there is a need for better fine-tuning to improve how LLMs deal with ambiguity. Second, LLMs should better adapt to different user needs. It is especially concerning that a request for simpler language leads to less thoughtful responses and fewer clarifications, showing that current systems often fail to support users with varied communication styles.

To support reproducibility and future research, we release our code\footnote{\href{https://github.com/lukasellinger/itdepends}{https://github.com/lukasellinger/itdepends}}. Further links to models and datasets are provided in the repository.

\section*{Limitations}
\paragraph{Multilingual Scope and Dataset Size.}
Our study focuses on English, French, Russian, Arabic, and Chinese. For non-English languages, we relied on direct translations from English using automated tools, which can introduce translation bias, cultural mismatches, or loss of nuance. Future work should create native datasets for each language to ensure more accurate and culturally appropriate evaluation. Additionally, the ClearRef and SharedRef datasets contain only 52 and 227 datapoints, respectively, and include only 8 relations from ConceptNet, making it difficult to draw fully stable conclusions and potentially biasing evaluation toward certain categories. Nevertheless, we observe very strong tendencies in the results, suggesting that the findings are still meaningful and indicative of broader trends.

\paragraph{Referential Order.} 
Due to computational limits, we used a fixed entity order; full permutation results for English are provided in \autoref{app:abl-permutation}.

\paragraph{Commonsense Context.}
We provided all models with the same context, which included a commonsense fact sourced from ConceptNet. While these facts consist of basic relations and vocabulary, we cannot guarantee that models internally represent or utilize this knowledge. Nevertheless, given the simplicity and generality of the facts, the models likely have access to such information.

\paragraph{LLM-based Evaluation.}
We used an LLM to judge model responses, observing near-perfect agreement with human annotations in English. While we did not conduct human agreement checks for other languages, the observed trends remain consistent across all languages, suggesting broader applicability. Moreover, the differences between prompt settings are substantially larger than any potential error margin, further reinforcing the robustness of our findings.

\paragraph{Selected Prompts.}
We use fixed user prompts for each relation, along with a single predefined suffix for requesting responses in simplified language. This setup reflects how typical users might interact with a model without actively optimizing prompt phrasing. However, LLMs are known to be highly sensitive to prompt formulation, which can significantly influence output quality~\cite{brown_language_2020}. Future research could systematically investigate the effects of varied or optimized prompts on LLM performance.

%
%

\bibliography{main}

\appendix
\section{Model Access}
\label{app:model_access}
To support reproducibility, \autoref{tab:models} lists all models used in this paper, including their abbreviated names (as used in tables and figures), full names, versions, and access providers.

\section{Dataset}
\label{app:dataset}
We extracted entities from the following eight relations: \texttt{CapableOf fly}, \texttt{HasProperty sweet}, \texttt{MadeOf wood}, \texttt{CapableOf swim}, \texttt{CapableOf run\_fast}, \texttt{CapableOf climb\_trees}, \texttt{HasProperty hot}, and \texttt{HasProperty loud}. All entities were manually reviewed and cleaned. During dataset construction, we used the following prompt with GPT-4.1-nano to verify that each negative entity truly does not satisfy the relation, in contrast to the two positive entities:
\begin{tcolorbox}[
  colback=bluebg,
  colframe=blue!60!black,
  coltitle=white,
  fonttitle=\bfseries,
  fontupper=\small\ttfamily,
  boxrule=0.5mm,
  rounded corners,
  title={User Prompt: Relation Satisfaction}
]
Does the word '<word>' satisfy the relation '<relation>'?\\ 
Answer with a brief explanation and either True or False for satisfies.
\end{tcolorbox}

\section{Ablation: Chain-of-Thought Prompting}
\label{app:abl-cot-prompt}
\citet{bian_chatgpt_2024} observe that LLMs often fail to apply commonsense knowledge when answering questions that require such reasoning. To investigate this in our setting, we tested GPT-4o on the English SharedRef dataset in a Chain-of-Thought (CoT) setting. We choose GPT-4o as it showed the sharpest drop from Normal to Simple. We appended the following instructions to encourage CoT reasoning:

\begin{tcolorbox}[
  colback=bluebg,
  colframe=blue!60!black,
  coltitle=white,
  fonttitle=\bfseries,
  fontupper=\small\ttfamily,
  boxrule=0.5mm,
  rounded corners,
  title={User Prompt: Chain-of-Thought}
]
<question> First, try resolving any ambiguity using commonsense knowledge. If the question remains ambiguous, your answer should be a clarification request. Otherwise, provide the answer. Put your final response after Response:.
\end{tcolorbox}

\begin{figure}[t]
    \centering

    \begin{minipage}{0.455\columnwidth}
        \centering
        \includegraphics[width=\linewidth]{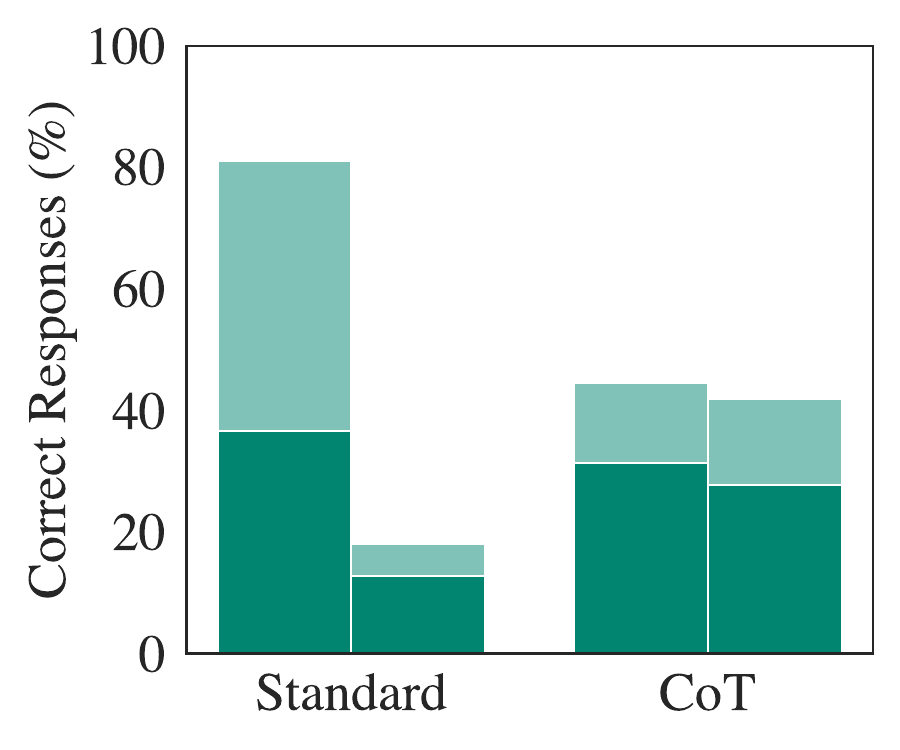}
    \end{minipage}
    \begin{minipage}{0.455\columnwidth}
        \centering
        \includegraphics[width=\linewidth]{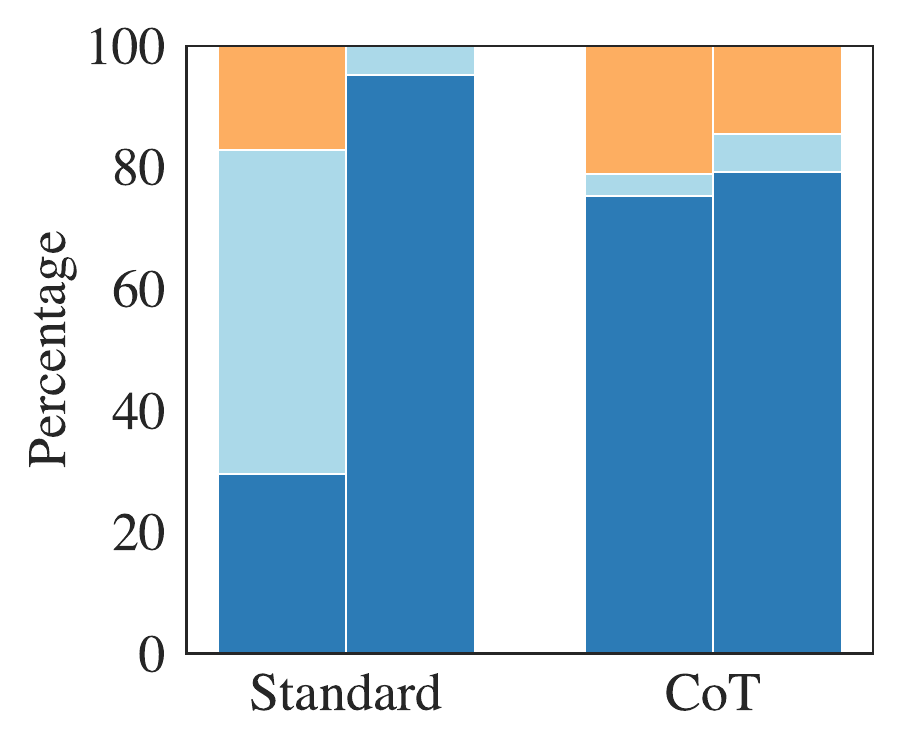}
    \end{minipage}
\caption{
Comparison of Standard vs. CoT-Prompting on the SharedRef dataset. 
Left: Correctness; the darker portion of each bar indicates the percentage of Direct Responses. 
Right: Response category distribution.
(Normal = left bar, Simple = right bar). 
Categories: 
\textcolor{answer_attempt}{\rule{1.2ex}{1.2ex}}~Answer Attempt, 
\textcolor{hedge}{\rule{1.2ex}{1.2ex}}~Hedge, 
\textcolor{clarification}{\rule{1.2ex}{1.2ex}}~Clarification, 
}
    \label{fig:cot}
\end{figure}

We compare standard and CoT prompting in \autoref{fig:cot}. CoT prompting performs worse than standard prompting, with accuracy dropping from 81.06\% to 44.49\% in the Normal setting. This is because CoT prompting often only partially resolves the ambiguity, responding to one positive while ignoring the other. This occurs roughly 50\% of the time, suggesting a model preference for one entity, as it correctly identifies each entity when prompted individually. We observe more Clarifications and Answer Attempts, with nearly no Hedging in the Normal setting. The Simple setting is largely similar, contrasting with the standard Simple prompting.

Comparing the gap between Normal and Simple settings, we find it much smaller than in standard prompting. This suggests that when the LLM is explicitly guided on how to generate responses, there is no loss of thoughtfulness or omission of important distinctions. This is also reflected in the Simple CoT setting, performing better than the Simple standard prompting.

\section{Ablation: Permutation of Entity Ordering}
\label{app:abl-permutation}

\begin{figure}[th]
    \centering

    \begin{minipage}{0.455\columnwidth}
        \centering
        \includegraphics[width=\linewidth]{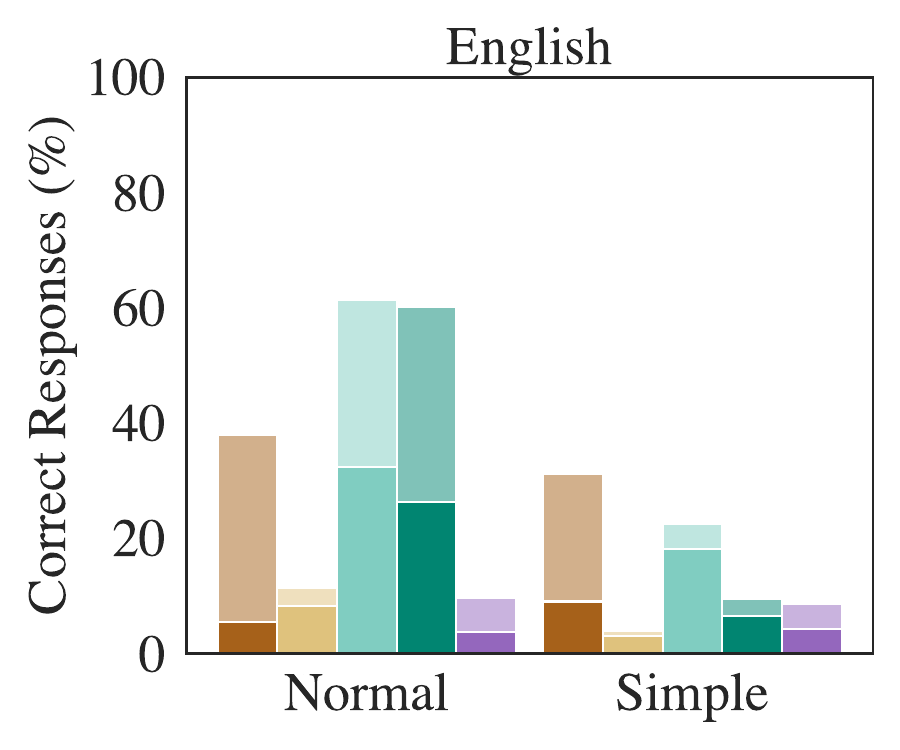}
    \end{minipage}
    \begin{minipage}{0.455\columnwidth}
        \centering
        \includegraphics[width=\linewidth]{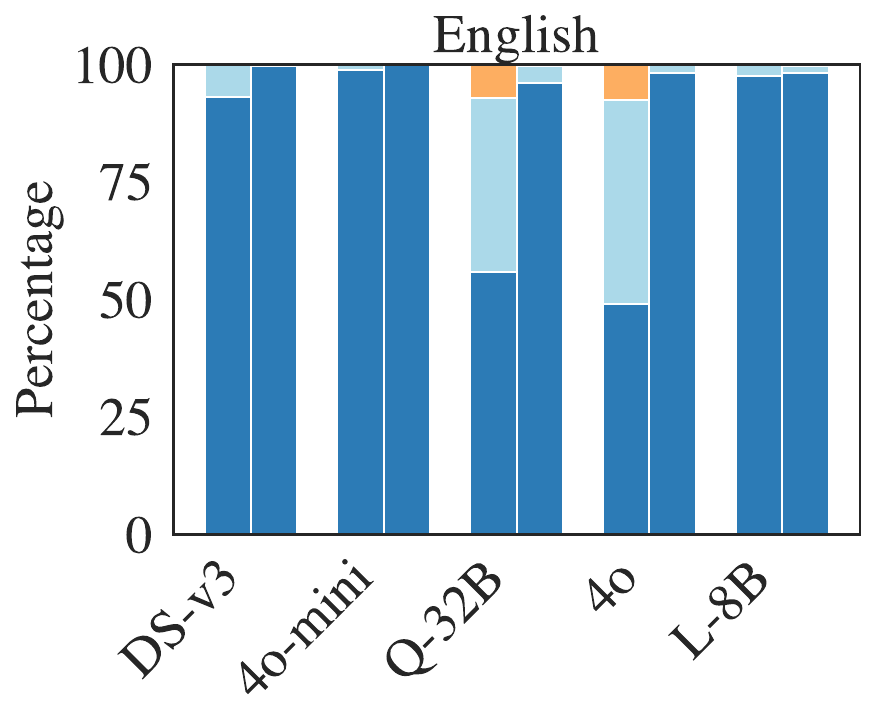}
    \end{minipage}
\caption{
Average performance across all permutations in the English SharedRef dataset. 
Left: Correctness per model; the darker portion of each bar indicates the percentage of Direct Responses. 
Right: Response category distribution (Normal = left bar, Simple = right bar). 
Models: 
\textcolor{deepseek}{\rule{1.2ex}{1.2ex}}~DeepSeek v3, 
\textcolor{gptmini}{\rule{1.2ex}{1.2ex}}~GPT-4o-mini, 
\textcolor{qwen}{\rule{1.2ex}{1.2ex}}~Qwen3-32B, 
\textcolor{gptfour}{\rule{1.2ex}{1.2ex}}~GPT-4o, 
\textcolor{llama}{\rule{1.2ex}{1.2ex}}~Llama-3.1-8B. 
Categories: 
\textcolor{answer_attempt}{\rule{1.2ex}{1.2ex}}~Answer Attempt, 
\textcolor{hedge}{\rule{1.2ex}{1.2ex}}~Hedge, 
\textcolor{clarification}{\rule{1.2ex}{1.2ex}}~Clarification, 
\textcolor{refuse}{\rule{1.2ex}{1.2ex}}~Refuse.
}
    \label{fig:avg-shared_ref}
\end{figure}

\begin{table}[th]
\centering
\begin{tabular}{@{}lccc@{}}
\toprule
\textbf{Prompt / Model} & \textbf{Pos. 1} & \textbf{Pos. 2} & \textbf{Pos. 3} \\
\midrule
\multicolumn{4}{l}{\textbf{Prompt: Normal}} \\%
\openai{} 4o & 34.50 & 24.52 & 40.98 \\
\openai{} 4o-mini & 41.98 & 18.65 & 39.37 \\
\deepseek{} v3 & 29.13 & 26.23 & 44.64 \\
\qwen{} 3-32B & 31.23 & 29.37 & 39.40 \\
\meta{} 3.1-8B & 35.93 & 17.41 & 46.66 \\
DPO Llama (Ours) & 33.38 & 32.34 & 34.28  \\
\midrule
\multicolumn{4}{l}{\textbf{Prompt: Simple}} \\%
\openai{} 4o & 37.29 & 15.80 & 46.91 \\
\openai{} 4o-mini & 41.67 & 16.46 & 41.87 \\
\deepseek{} v3 & 31.03 & 25.75 & 43.23 \\
\qwen{} 3-32B & 30.35 & 28.36 & 41.29 \\
\meta{} 3.1-8B & 35.04 & 16.53 & 48.43 \\
DPO Llama (Ours) & 33.73 & 32.30 & 33.97 \\
\bottomrule
\end{tabular}
\caption{Average selection rate (\%) of an entity appearing at Position 1, 2, or 3 in the SharedRef dataset, across different models and prompts (Normal vs. Simple) in English.}
\label{tab:ablation-pos-sharedref}
\end{table}

Our conversation context has a given order of entities. Due to computational constraints, we fixed the order to a single permutation for all evaluations (`0, 1, 2' for SharedRef and `0, 1' for ClearRef). We based this choice not on performance but to ensure consistency across languages.

To assess the effect of this choice, we ran an ablation on the English dataset using all permutations. We observed that the frequency with which a model selects an entity depends heavily on its position in the list, indicating a strong positional bias.

\autoref{tab:ablation-pos-sharedref} shows the distribution of selected entities across positions for each permutation in SharedRef. For example, in the Simple setting, entities at position three are selected drastically more often (avg. 42.62\%) than those at position two (avg. 22.53\%).

\autoref{tab:ablation-pos-clearref} presents analogous results for ClearRef. Here, the bias is milder, with position two being selected slightly more frequently on average (+4.22\% in Normal, +3.03\% in Simple).

\begin{table}[t]
\centering
\begin{tabular}{@{}lcc@{}}
\toprule
\textbf{Prompt / Model} & \textbf{Pos. 1} & \textbf{Pos. 2} \\
\midrule
\multicolumn{3}{l}{\textbf{Prompt: Normal}} \\%
\openai{} 4o & 48.00 & 52.00\\
\openai{} 4o-mini & 48.35 & 51.65\\
\deepseek{} v3 & 43.36 & 56.64\\
\qwen{} 3-32B & 48.15 & 51.85\\
\meta{} 3.1-8B & 48.57 & 51.43\\
DPO Llama (Ours) & 50.90& 49.10\\
\midrule
\multicolumn{3}{l}{\textbf{Prompt: Simple}} \\%
\openai{} 4o & 49.48 & 50.52 \\
\openai{} 4o-mini & 48.65 & 51.35\\ 
\deepseek{} v3 & 48.04 & 51.96\\
\qwen{} 3-32B & 46.24 & 53.76\\
\meta{} 3.1-8B & 48.51 & 51.49\\
DPO Llama (Ours) & 50.00 & 50.00 \\
\bottomrule
\end{tabular}
\caption{Average selection rate (\%) of an entity appearing at Position 1, or 2 in the ClearRef dataset, across different models and prompts (Normal vs. Simple) in English.}
\label{tab:ablation-pos-clearref}
\end{table}

\autoref{fig:avg-shared_ref} shows the averaged correctness and category distribution over all permutations in English SharedRef. Compared to the fixed `0, 1, 2' ordering used in our main results, average correctness drops. Notably, GPT-4o exhibits fewer clarification attempts when averaged across permutations, while Qwen3-32B maintains strong performance. 

The overall trend of higher correctness and better category distribution in the Normal setting compared to the Simple setting remains.

\section{Response Categorization}
\label{app:response-categories}
We adopt the response taxonomy proposed by \citet{herlihy_overcoming_2024}, with slight modifications to better suit our coreference resolution evaluation. Specifically, we merge \textit{Interrogation} into \textit{Clarification}, as both involve follow-up questions rather than direct answers. We also merge \textit{Discussion} into \textit{Answer Attempt}, since our evaluation does not require a fully factual answer, only that the response correctly identifies the positive entities. This simplification reduces annotation complexity without compromising the core objective of our analysis. A detailed overview of all response categories, including definitions and examples, is provided in \autoref{tab:response_types}.

\begin{table*}[ht]
    \centering
    \begin{tabular}{lp{6cm}p{6cm}}
        \toprule
        \textbf{Name} & \textbf{Description} & \textbf{Example} \\
        \midrule
        Answer attempt & The assistant clearly commits to at least one interpretation, providing a factual response without any hedging. & The helicopter can fly because its rotors generate lift, allowing it to rise off the ground and maneuver through the air.\\
        \midrule
        Clarification & The response asks for more information about the user’s intent without offering interpretations or using hedging & Could you specify which one you're referring to: the helicopter or the gnat? \\
        \midrule
        Hedging & The response uses hedging or conditional language (e.g., ``if you meant X...'') and does not fully commit to a single interpretation. Even if only one entity is mentioned, the presence of such language marks it as a hedge. & If you're referring to a raven, it can fly due to its strong wings, lightweight body, and aerodynamic shape, which allow it to generate lift and move through the air efficiently.\\
        \midrule
        Refuse & The response refuses to answer the question and does not ask any follow-up questions. & Here are the sentences: 1. **Bat**: The bat fluttered silently through the night sky, searching for insects. 2. **Dragonfly**: The dragonfly darted over the pond, its iridescent wings shimmering in the sunlight. 3. **Coffee**: The aroma of freshly brewed coffee filled the kitchen, awakening everyone’s senses. 4. **Why can it fly?**: Why can it fly, despite its small wings and heavy body?" \\
        \midrule
        Missing & The response is empty. & [blank] \\
        \bottomrule
    \end{tabular}
    \caption{Description and Examples of our Response Categories.}
    \label{tab:response_types}
\end{table*}

\section{Automatic Evaluation}
\label{app:automatic-evaluation}
We used GPT-4.1-mini as an LLM judge to automatically evaluate the responses. We divided the evaluation into two parts: response classification and entity extraction. The prompt used for response classification is shown in \autoref{box:response-classification}. The prompts used for entity extraction, split into a system prompt and a user prompt, are shown in \autoref{box:sys-entity} and \autoref{box:user-entity}, respectively.

We manually annotated 500 responses from the English dataset to validate the framework. \autoref{tab:human-eval} reports the agreement rates for response categorization along with Cohen’s Kappa scores. For entity extraction, we report exact match accuracy. Overall, the results show high agreement across all models.

\begin{table}[ht]
\centering
\begin{tabular}{lcc}
\toprule
\textbf{Prompt} & \textbf{Response Cat.} & \textbf{Entity} \\
\midrule
\openai{} 4o-mini   & 100.0\% (N/A)  & 99\% \\
\openai{} 4o & 100.0\% (1.000) & 98\% \\
\qwen{} 3-32B & 92.0\% (0.804) & 98\% \\
\deepseek{} v3      & 98.0\% (0.823) & 94\% \\
\meta{} 3.1-8B  & 100.0\% (N/A) & 100\% \\
\textbf{Total} & 98.0\% (0.916) & 97.8\% \\
\bottomrule
\end{tabular}
\caption{Accuracy percentages and Cohen’s Kappa scores (in parentheses) for Response Categorization and exact match accuracy for Entity Extraction across our evaluated models.}
\label{tab:human-eval}
\end{table}

\section{Direct Preference Optimization}
\label{app:dpo}
Our training set contains 472 responses from simple settings and 866 from normal settings. In addition, we included 30 basic clarification cases, where the user posed clearly ambiguous questions. A fine-grained distribution is provided in \autoref{tab:dpo-dataset}.

\begin{table}[t]%
\centering%
\small%
\begin{tabular}{@{}lrrrrr@{}}%
\toprule%
\textbf{Dataset / Category} &En&Fr&Ar&Ru&Zh\\%
\midrule%
\multicolumn{6}{l}{\textbf{SharedRef}} \\%
Normal Answer Attempt  &64   &80 &69 &37 &53\\%
Normal Hedge           &106  &39 &49 &78 &57\\%
Normal Clarification   &58   &44 &47 &55 &47\\%
Simple Answer Attempt  &112&84 &30 &69&76\\%
Simple Hedge           &21 &13 &2  &15&31\\%
Simple Clarification   &4  &3  &1  &4 &1\\%
\midrule%
\multicolumn{6}{l}{\textbf{ClearRef}} \\%
Normal Answer Attempt&&&2&&\\%
Normal Hedge&&&1&&\\%
Simple Answer Attempt&&&6&&\\%
\midrule%
\multicolumn{6}{l}{\textbf{General}} \\%
Clarification&6&6&6&6&6\\\bottomrule%
\end{tabular}%
\caption{Distribution of chosen response types in our DPO fine-tuning dataset, broken down by language, response category, and setting.}
\label{tab:dpo-dataset}
\end{table}

We fine-tuned the model for two epochs using Low-Rank Adaptation (LoRA). The full configuration for LoRA and DPO training is summarized in \autoref{tab:dpo-finetuning-specs}.

\begin{table}[t]
\centering
\begin{tabular}{ll}
\toprule
\textbf{Parameter} & \textbf{Value} \\
\midrule
\multicolumn{2}{l}{\textit{LoRA Configuration}} \\
$r$ & 64 \\
LoRA Alpha & 16 \\
LoRA Dropout & 0.05 \\
Target Modules & \makecell[l]{[q\_proj, v\_proj,\\ k\_proj, o\_proj]} \\
Bias & none \\
\midrule
\multicolumn{2}{l}{\textit{DPO Training Configuration}} \\
$\beta$ & 0.1 \\
Learning Rate & 5e-5 \\
Batch Size (per device) & 4 \\
Epochs & 2 \\
\bottomrule
\end{tabular}
\caption{Combined configuration used for LoRA adaptation and Direct Preference Optimization (DPO) fine-tuning.}
\label{tab:dpo-finetuning-specs}
\end{table}

\begin{figure}[t]
    \centering

    \includegraphics[width=\linewidth]{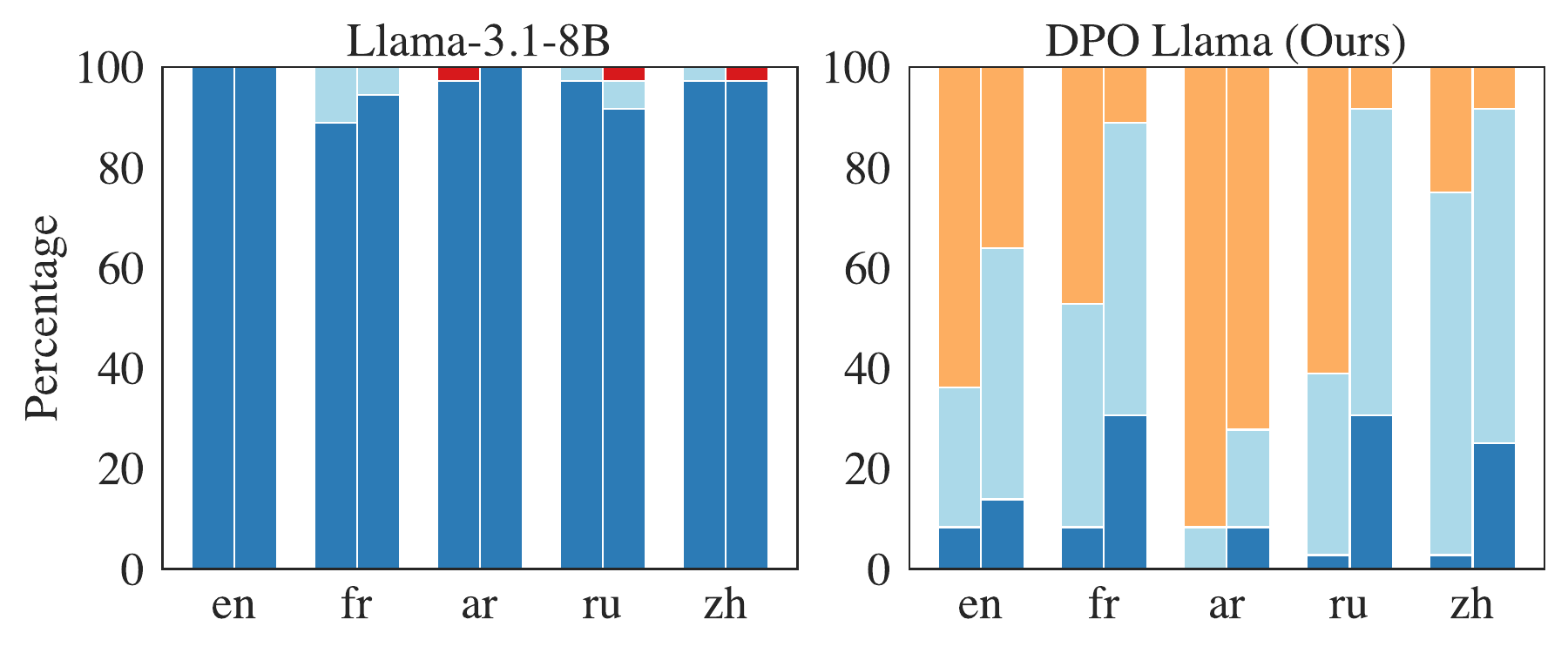}
    \caption{
    Distribution of response categories in the DPO test set across five languages in the ClearRef dataset. 
    Colored squares denote response types: 
    \textcolor{answer_attempt}{\rule{1.2ex}{1.2ex}}~Answer Attempt, 
    \textcolor{hedge}{\rule{1.2ex}{1.2ex}}~Hedge, 
    \textcolor{clarification}{\rule{1.2ex}{1.2ex}}~Clarification, and 
    \textcolor{refuse}{\rule{1.2ex}{1.2ex}}~Refuse.
    }
    \label{fig:dpo-clear-ref}
\end{figure}

We observed performance improvements on both the SharedRef dataset and the homonym task from \citet{ellinger_simplifications_2025}. However, on the ClearRef test set, while the number of correct responses remained comparable to the base model, we experienced a category shift. As shown in \autoref{fig:dpo-clear-ref}, the distribution of coarse response categories shifted significantly toward `clarification' and `hedge' across all languages. This indicates that the cognitive cost of those responses is higher for our DPO model compared to the base model on this dataset. To address this, future alignment efforts should incorporate more training examples from ClearRef to encourage direct answers where appropriate. Unlike in SharedRef, where the model successfully used common knowledge to respond only to the positive entities, in ClearRef, the model no longer consistently applies this strategy.\newpage

\begin{table*}[t]%
\centering%
\begin{tabular}{@{}l l l l@{}}%
\toprule%
Short Form&Name&Version&Access Provider\\%
\midrule%
\openai{} 4o-mini   &GPT-4o-mini&gpt-4o-mini-2024-07-18&OpenAI API\\%
\openai{} 4o        &GPT-4o     &gpt-4o-2024-08-06&OpenAI API\\%
\openai{} 4.1-nano  &GPT-4.1-nano     &gpt-4.1-nano-2025-04-14&OpenAI API\\%
\openai{} 4.1-mini  &GPT-4.1-mini    &gpt-4.1-mini-2025-04-14&OpenAI API\\%
\qwen{} 3-32B       &Qwen3-32B  &N/A&OpenRouter\\%
\deepseek{} v3      &Deepseek v3&N/A&Fireworks AI\\%
\meta{} 3.1-8B      &Llama-3.1-8B&N/A&Fireworks AI\\\bottomrule%
\end{tabular}%
\caption{Specific model versions used in our experiments. For each model we provide the short form as used in our tables, the exact version and the access provider.}%
\label{tab:models}
\end{table*}

\begin{table*}[h!]%
\centering%
\setlength{\tabcolsep}{5pt}
\begin{tabular}{@{}lrrrrrrrrrr@{}}%
\toprule%
\textbf{Prompt / Model} & \multicolumn{5}{c}{\textbf{Correct}} & \multicolumn{5}{c}{\textbf{Direct}}\\%
\cmidrule(lr){2%
-%
6}%
\cmidrule(lr){7%
-%
11}%
&En&Fr&Ar&Ru&Zh&En&Fr&Ar&Ru&Zh\\%
\midrule%
\multicolumn{11}{l}{\textbf{Prompt: Simple}} \\%
\openai{} 4o&98.08&\textbf{100.00}&98.08&\textbf{100.00}&96.15&88.46&69.23&\textbf{78.85}&63.46&75.00\\%
\openai{} 4o-mini&\textbf{100.00}&\textbf{100.00}&\textbf{100.00}&\textbf{100.00}&\textbf{100.00}&69.23&55.77&53.85&28.85&76.92\\%
\deepseek{} v3&\textbf{100.00}&90.38&\textbf{100.00}&94.23&\textbf{100.00}&88.46&67.31&61.54&69.23&69.23\\%
\qwen{} 3-32B&\textbf{100.00}&94.23&94.23&96.15&96.15&73.08&63.46&21.15&65.38&67.31\\%
\meta{} 3.1-8B&94.23&94.23&98.08&90.38&92.31&\textbf{92.31}&\textbf{78.85}&55.77&\textbf{73.08}&\textbf{80.77}\\%
\midrule%
\multicolumn{11}{l}{\textbf{Prompt: Normal}} \\%
\openai{} 4o&96.15&96.15&\textbf{98.08}&\textbf{98.08}&96.15&86.54&73.08&\textbf{75.00}&\textbf{80.77}&80.77\\%
\openai{} 4o-mini&\textbf{100.00}&\textbf{98.08}&94.23&\textbf{98.08}&94.23&82.69&65.38&57.69&46.15&67.31\\%
\deepseek{} v3&\textbf{100.00}&\textbf{98.08}&96.15&94.23&96.15&69.23&53.85&59.62&61.54&50.00\\%
\qwen{} 3-32B&98.08&94.23&\textbf{98.08}&90.38&94.23&82.69&51.92&51.92&67.31&63.46\\%
\meta{} 3.1-8B&96.15&90.38&90.38&94.23&\textbf{98.08}&\textbf{94.23}&\textbf{78.85}&65.38&73.08&\textbf{90.38}\\\bottomrule%
\end{tabular}%
\caption{Evaluation results showing the percentage of correct and direct responses across languages and prompt types on the ClearRef dataset. \textbf{Bold} highlights the highest scores per language within each prompt and metric.}%
\label{tab:clear_ref}
\end{table*}

\begin{table*}[h!]%
\centering%
\setlength{\tabcolsep}{5pt}%
\begin{tabular}{@{}lrrrrrrrrrr@{}}%
\toprule%
\textbf{Prompt / Model} & \multicolumn{5}{c}{\textbf{Correct}} & \multicolumn{5}{c}{\textbf{Direct}}\\%
\cmidrule(lr){2%
-%
6}%
\cmidrule(lr){7%
-%
11}%
&En&Fr&Ar&Ru&Zh&En&Fr&Ar&Ru&Zh\\%
\midrule%
\multicolumn{11}{l}{\textbf{Prompt: Simple}} \\%
\openai{} 4o&18.06&7.49&19.82&4.41&23.79&12.78&6.61&\textbf{12.78}&2.20&15.42\\%
\openai{} 4o-mini&5.29&1.76&11.45&3.52&11.89&3.52&1.76&8.81&1.76&8.37\\%
\deepseek{} v3&\textbf{48.02}&\textbf{28.19}&\textbf{47.58}&\textbf{19.38}&46.70&10.57&\textbf{11.89}&6.17&3.96&7.49\\%
\qwen{} 3-32B&33.04&25.99&28.19&12.33&\textbf{56.83}&\textbf{23.35}&\textbf{11.89}&3.96&\textbf{7.93}&\textbf{26.87}\\%
\meta{} 3.1-8B&11.89&3.96&\textbf{47.58}&4.85&4.85&4.85&0.44&2.20&1.76&1.76\\%
\midrule%
\multicolumn{11}{l}{\textbf{Prompt: Normal}} \\%
\openai{} 4o&\textbf{81.06}&50.66&47.14&57.27&62.11&\textbf{36.56}&\textbf{34.80}&\textbf{30.40}&22.03&21.15\\%
\openai{} 4o-mini&14.10&11.89&14.10&4.41&18.50&7.05&7.49&8.37&1.76&7.05\\%
\deepseek{} v3&61.23&61.23&\textbf{65.64}&31.28&49.78&3.96&9.25&10.13&3.52&1.76\\%
\qwen{} 3-32B&75.77&\textbf{71.37}&56.83&\textbf{63.88}&\textbf{83.26}&29.52&19.38&10.57&\textbf{25.99}&\textbf{25.11}\\%
\meta{} 3.1-8B&14.54&13.66&7.49&3.96&6.17&5.73&2.20&3.96&1.32&2.64\\\bottomrule%
\end{tabular}%
\caption{Evaluation results showing the percentage of correct and direct responses across languages and prompt types on the SharedRef dataset. \textbf{Bold} highlights the highest scores per language within each prompt and metric.}%
\label{tab:shared_ref}
\end{table*}

\begin{responsebox*}[label={box:response-classification}]{User Prompt: Response Classification}
\begin{verbatim}
You are reviewing a single-turn conversation between a user and an assistant.

You must classify the assistant's response according to the **type of response strategy** used.

Here are the definitions:

- `answer_attempt`: The assistant **unambiguously commits** to at least one interpretation of the 
user’s question and provides a factual explanation or general background relevant to the query. It 
must **not use any hedging language**, such as “if you mean…”, “depending on…”, or “could 
be…”. Implicit or explicit hedging disqualifies it from this category. It may mention multiple 
entities, as long as none are introduced with hedging or conditional phrasing.

- `clarification`: The assistant **asks for more information** about the user’s intent. This may 
involve **one or more clarification questions**, but must not **suggest answers** or list 
**interpretations**,. The response must should **not contain hedging or speculative language**.

- `hedge`: The assistant **uses hedging or conditional language** (e.g., “if you meant X…”, 
“could refer to…”, “might be…”) and does not fully commit to a single interpretation. Even if 
only one entity is mentioned, the presence of such language marks it as a hedge.
        
- `refuse`: The assistant explicitly or implicitly refuses to answer the question and does **not** 
ask any follow-up questions.

- `missing`: The assistant gives **no response at all**.

You must output your answer in the following JSON format:
{"response_type": "refuse|missing|answer_attempt|hedge|clarification"}

Conversation:
Question: <question_str>
Answer: <asnwer_str>
\end{verbatim}
\end{responsebox*}

\begin{responsebox*}[label={box:sys-entity}, colback=gray!20, colframe=gray!60!black]{System Prompt: Entity Extraction}
You are an expert in identifying explicit entity mentions in text.\\

Your task is to determine which of the listed entities are explicitly mentioned in the response.\\

Instructions:\\
- Return only entities from the provided list (use their exact spelling from the list in your output).\\
- Consider an entity 'mentioned' if the response text contains the entity exactly, or with minor morphological or spelling variations.\\
Examples of acceptable variations include plural forms and misspellings.\\
- Do not infer mentions from implied meaning, associations, or unrelated paraphrasing.\\
- Only include entities that are clearly and explicitly present in the response text.\\
- **Always** output the entity names exactly as they appear in the provided list.
\end{responsebox*}

\begin{responsebox*}[label={box:user-entity}]{User Prompt: Entity Extractionn}
Entities: <entity\_list>\\
Response:\\

<answer\_str>\\

Which of the listed entities are explicitly mentioned in the response?
\end{responsebox*}

\begin{figure*}[ht]
\centering
\pdfbox{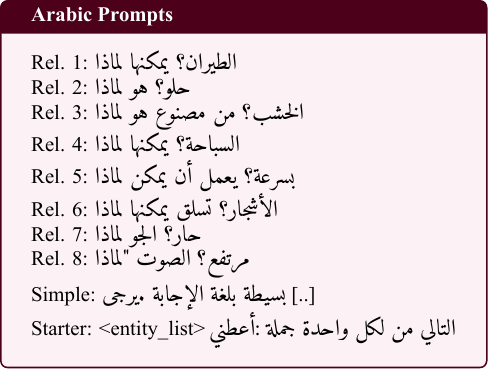}
\pdfbox{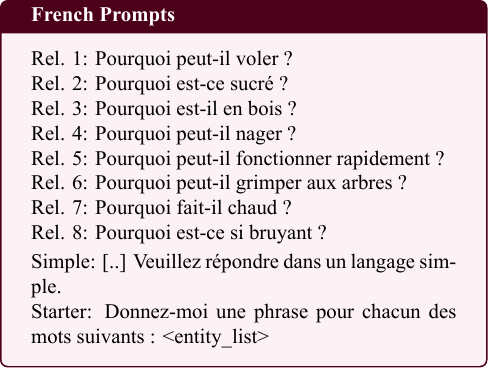}

\vspace{0.5em}

\pdfbox{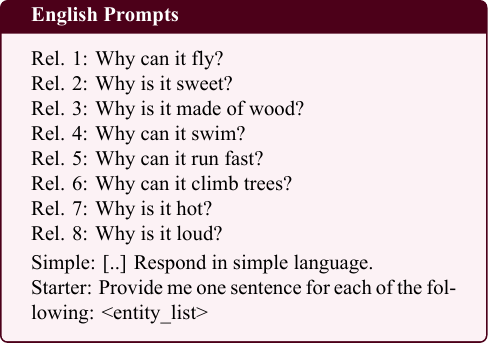}
\pdfbox{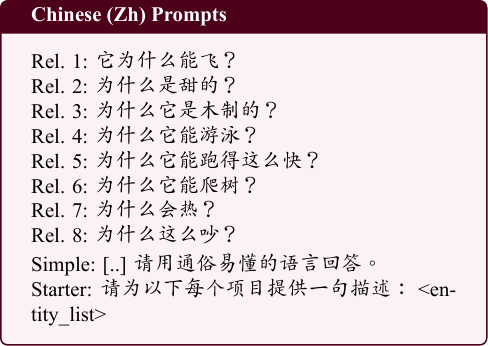}

\vspace{0.5em}

\pdfbox{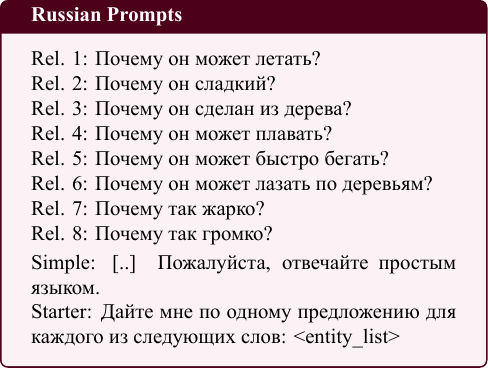}

\caption{Language Versions of Relation Questions, the Simple Instruction and the Starter Sentence in Arabic, French, English, Chinese, and Russian}
\label{fig:multilingual-prompts}
\end{figure*}

\end{document}